\documentclass[conference]{IEEEtran}
\IEEEoverridecommandlockouts
\usepackage{amsmath,amssymb,amsfonts}
\usepackage{graphicx}
\usepackage{textcomp}
\usepackage{xcolor}
\def\BibTeX{{\rm B\kern-.05em{\sc i\kern-.025em b}\kern-.08em
    T\kern-.1667em\lower.7ex\hbox{E}\kern-.125emX}}
\DeclareMathAlphabet\mathbfcal{OMS}{cmsy}{b}{n}
\usepackage[utf8]{inputenc}
\usepackage{graphicx}
\graphicspath{ {./images/} }
\usepackage{color}
\usepackage[x11names, table]{xcolor}
\usepackage[dvipsnames]{xcolor}
\usepackage[T1]{fontenc}

\newcommand*{\dittostraight}{---\textquotedbl---} 
\usepackage{algorithm}
\usepackage{algpseudocode}
\usepackage{graphicx}
\usepackage{booktabs}
\usepackage{longtable}
\usepackage{caption}
\usepackage{subcaption}
\usepackage{tabularx}
\usepackage{array}
\usepackage{ragged2e}
\usepackage{amsmath}
\newcommand{\thickbar}[1]{\mathbf{\bar{\text{$#1$}}}}
\usepackage{pifont}

\usepackage{rotating}
\usepackage{hyperref}
\usepackage{hyperxmp}
\usepackage{multirow}
\usepackage{stfloats}
\usepackage{orcidlink}
\newcommand{\fastman}{{\fontfamily{lmtt}\selectfont FaSTM$\forall$N}}
\newcommand{\redirect}{{\fontfamily{lmtt}\selectfont ReDiRect}}
\usepackage[
backend=biber,
style=numeric,
sorting=ynt
]{biblatex}
\PassOptionsToPackage{hyphens}{url}
\usepackage{tikz}
\def\checkmark{\tikz\fill[scale=0.4](0,.35) -- (.25,0) -- (1,.7) -- (.25,.15) -- cycle;} 
\usepackage{wrapfig}
\usepackage{float}

\begin{document}
\title{Detecting Complex Money Laundering Patterns with Incremental and Distributed Graph Modeling}

\author{\IEEEauthorblockN{Haseeb Tariq~\orcidlink{0000-0003-0756-3714}}
\IEEEauthorblockA{\textit{ING Bank} \\
Amsterdam, The Netherlands \\
\textit{Eindhoven University of Technology}\\
Eindhoven, The Netherlands \\
m.h.tariq@tue.nl}
\and
\IEEEauthorblockN{Alen Kaja~\orcidlink{0009-0001-9793-3813}}
\IEEEauthorblockA{\textit{Identity \& Risk Intelligence} \\
\textit{Adyen}\\
Amsterdam, The Netherlands \\
alen.kaja@adyen.com}
\and
\IEEEauthorblockN{Marwan Hassani~\orcidlink{0000-0002-4027-4351}}
\IEEEauthorblockA{\textit{Mathematics \& Computer Science} \\
\textit{Eindhoven University of Technology}\\
Eindhoven, The Netherlands \\
m.hassani@tue.nl}
}

\maketitle

\begin{abstract}
Money launderers take advantage of limitations in existing detection approaches by hiding their financial footprints in a deceitful manner. They manage this by replicating transaction patterns that the monitoring systems cannot easily distinguish. As a result, criminally gained assets are pushed into legitimate financial channels without drawing attention. Algorithms developed to monitor money flows often struggle with scale and complexity. The difficulty of identifying such activities is further intensified by the (persistent) inability of current solutions to control the excessive number of false positive signals produced by rigid, risk-based rules systems. We propose a framework called {\redirect} (\underline{Re}duce, \underline{Di}stribute, and \underline{Rect}ify), specifically designed to overcome these challenges. The primary contribution of our work is a novel framing of this problem in an unsupervised setting; where a large transaction graph is \textit{fuzzily} partitioned into smaller, manageable components to enable fast processing in a distributed manner. In addition, we define a refined evaluation metric that better captures the effectiveness of exposed money laundering patterns. Through comprehensive experimentation, we demonstrate that our framework achieves superior performance compared to existing and state-of-the-art techniques, particularly in terms of efficiency and real-world applicability. For validation, we used the \textit{real} (open source) Libra dataset %\cite{ego2}. We also used 
and the recently released synthetic datasets by IBM Watson.
%\cite{synthdata} as the \textit{secondary} source for benchmarking
% Furthermore, our work includes a critical evaluation of the practical relevance and limitations of these synthetic datasets. 
Our code and datasets are
available at https://github.com/mhaseebtariq/redirect.
\end{abstract}


\section{Introduction}
\label{sec:introduction}

The United Nations Office on Drugs and Crime (UNODC) has estimated that approximately 2 to 5\% of global GDP is laundered annually, which translates to between 715 billion and 1.87 trillion Euros each year \cite{mloverview}. Quantifying the toll of human suffering is even more complex. Current financial monitoring systems have struggled to detect these laundering activities effectively, leading to significant penalties being imposed on major banks by the regulators. A key issue for banks is that they can only observe transactions within their own systems, while criminals typically move funds across multiple banks in different jurisdictions to conceal the complete trail of money.

The money laundering process is typically divided into three phases: Placement, Layering, and Integration \cite{mloverview}. In the placement phase, criminals introduce illicit funds into the financial system, often by depositing them in a manner that does not raise suspicion. In the subsequent layering phase, they convolute the money trail by dispersing funds across various intermediaries. Finally, in the integration phase, the money is withdrawn from accounts that appear to be operating under legitimate conditions. The methods criminals employ in these phases are referred to as \textit{typologies}. There are numerous well-documented typologies, such as \textit{smurfing} (or \textit{structuring}) \cite{smurfing}, where large sums of illicit funds are broken into smaller and less noticeable deposits in several \textit{mule accounts} \cite{mule}.

Criminals continuously develop new techniques (or typologies) to circumvent financial monitoring systems. This constant innovation makes it challenging for \textit{Anti-Money Laundering (AML)} experts to keep up with emerging risks. Financial institutions use a combination of \textit{Transaction Monitoring (TM)} systems and \textit{Know Your Customer (KYC)} protocols
% \cite{kyc}
to prevent money laundering and terrorist financing. They also implement frameworks such \textit{as Customer Due Diligence (CDD)} and \textit{Ultimate Beneficial Owner (UBO)} identification, in addition to rules-based frameworks, to identify high-risk transactions. However, a significant proportion of money laundering activities remain undetected because experts spend an overwhelming amount of time sifting through and discarding a massive number of \textit{false positive alerts}. In practice, this represents the most substantial challenge for financial institutions. Banks spend tens of millions of Euros annually on processing these alerts. Otherwise, due to lack of regulatory compliance, fines can amount to hundreds of millions of Euros \cite{abnfine}. Reports on \textit{AML compliance costs} \cite{amlstats} provide staggering figures on the monetary burden faced by financial institutions.

Our proposed framework addresses these challenges by focusing on building more concise, fuzzy, and overlapping communities with the purpose of reducing the cognitive load for AML analysts. Consequently, (also) minimizing the \textit{false positive rate}. Our key contributions are as follows:
\begin{itemize}
\item A novel and scalable way to formulate the AML modeling problem
\item A data scope reduction technique that, in addition to scalability, also improves the model accuracy; and usefulness by \textit{drastically} reducing the ILT (Investigation Lead Time) of the (outputted) alerts 
\item A novel metric to measure the holistic context of AML anomalies
\end{itemize}

Section \ref{sec:related} discusses existing work. Sections \ref{sec:preliminaries} \& \ref{sec:method} introduce the datasets, definitions, and {\redirect}. Section \ref{sec:experiments} details the experimental evaluation; and Section \ref{sec:conclusion} concludes the paper.
\section{Related Work}
\label{sec:related}
AML modeling and fraud detection have several similarities; and face some common challenges. However, in the context of our research, we want to distinguish the two problem spaces. The focus of our research is strictly on detecting complex money laundering networks. More precisely, we want to identify how the anomalous behaviors of a money launderer are manifested as their transaction activities. We will briefly review some of the previous research on the detection of money laundering networks (and patterns) backed by graph algorithms. 

To our knowledge, with the exception of {\fastman} \cite{fastman}, there is no existing work that focuses on \textit{first} detecting communities (built on a transaction graph) with the purpose of identifying suspicious money laundering patterns in them. For example, FlowScope \cite{flowscope} and CubeFlow \cite{cubeflow} \textit{directly} identify suspicious flows in a global transaction graph; Graphomaly \cite{ego2} and MonLad \cite{monlad} identify suspicious \textit{nodes} by constructing graph features; and GraphFeatureProcessor \cite{graphfep} introduces a library to construct graph features with the purpose of identifying anomalous \textit{edges} (or transactions). In ExSTraQt \cite{exstraqt}, a similar but much more effective framework is proposed.

The modeling of transaction data as graphs is a popular choice \cite{amlgraph} for building AML models. \cite{study} presents a comparative study on some of the well-known community detection methods. The Louvain algorithm \cite{Blondel_2008} is one of the most widely used modularity-based community detection methods \cite{modularity}. An advanced version, called the Leiden algorithm, has been proposed in \cite{leiden}. Modularity-based algorithms produce non-overlapping communities. This means that a node can belong only to a single community. In the real world, this is not really the case for many problems. A node (or a business entity) can belong to several communities. Similarly, a money laundering agent can also support multiple criminal networks. That is why our framework also employs a bottom-up approach. Where, for each node, we build a community around it using the \textit{personalized} pagerank algorithm \cite{pagerank, fastppr}. This results in fuzzy communities that overlap. The study in \cite{wagenseller2017size} assesses various algorithms by analyzing the distributions of (resulting) community sizes. Anomalous communities eventually have to be analyzed by AML analysts. As the size of a community becomes too large, it becomes unrealistic to analyze it by \textit{observing} signs of suspicious patterns. One of our \textit{core} contributions focuses on reducing the size of the alerted (or outputted) communities. 

A detailed review published in \cite{akoglu2014graphbased} examines anomaly detection techniques for data represented as graphs. Most community detection methods are designed to identify communities with strong internal connections, while having weak external links to the \textit{other} communities. In transaction datasets (with low degrees of separation), this typically results in communities with short paths, often with just 1 or 2 hops. Consequently, simpler methods such as those in \cite{ego2} or \cite{oddball} are more effective in detecting anomalous communities with closely connected neighbors. Due to stringent oversight by data and financial regulators, AML models must be free of bias, restricting the number of usable data points. This includes, for example, avoiding gender, age, racial, socioeconomic, or geographic biases. Modelers must also be cautious of the \textit{guilt by association} fallacy. For example, the method proposed in \cite{supervised} risks falling into this trap. Data sharing among banks remains a risk, as privacy enhancing technologies such as multiparty computation and homomorphic encryption \cite{pet} are not mature enough to be used on a large scale; or for more practical applications. This pushes the need for improving the state-of-the-art for single-bank AML modeling.

Lastly, our research focuses on detecting the anomalous \textit{context} in which money laundering occurs. We use IsolationForest \cite{iforest} to detect anomalous nodes (or accounts). The detection method can be easily swapped with any graph-based neural network \cite{gnn} such as GraphSAGE \cite{graphsage} or VGAE \cite{vgae}. Therefore, we do not care much about the specific anomaly detection algorithm that is being used. More importantly, in a highly regulated industry like financial services, model interpretability is of the utmost importance. The output of the \textit{tabular} feature-based model we train in our framework can be easily explained by either 1) estimating a simple (linear) model on top of the predicted binary label (anomalous/normal); or 2) using a model interpretability framework like SHAP \cite{shap} or LIME \cite{lime}. There are no straightforward methods for using the same interpretability frameworks on (advanced) neural network architectures designed for graphical data.
\section{Datasets}
\label{sec:preliminaries}
Despite having access to real transaction data from some of the largest European banks, for this paper we will rely on open source datasets. This is because of the obvious risk implications that come with publishing results from real data, containing real money laundering cases. We built our framework using the \textit{real} Libra Internet Bank dataset, which we will call $\mathcal{D}_{lib}^{real}$. The content of this dataset is described in \cite{ego2}. 

We also used a synthetic dataset \cite{synthdata} published by the IBM Watson research lab in 2023 as a \textit{secondary} source. We are going to call it $\mathcal{D}_{ibm}^{syn}$. The content of this dataset is elaborately explained in \cite{synthdata}. There are 6 different versions of the dataset, with 2 categories. The first category represents the size of the dataset as low (1 month), medium (3 months), and large (6 months). The second category represents the ratio of anomalies (or illicit transactions) present in the data, as low-illicit and high-illicit. The dataset is injected with a rich set of labels on a transaction level; and on a (broader) \textit{pattern} level. The injected patterns can actually be seen as the anomalous context that we want to identify. The concept of \textit{context} is extensively explained in Section \ref{sec:eval-metric}, with the support of Figure \ref{fig:cont_metric} and Algorithm \ref{alg:wcm}. In our work, we focus mainly on large versions of the dataset. By doing so, we also cover the scalability aspects of our framework.

\section{Our Method: {\redirect}}
\label{sec:method}
\begin{figure*}[tp!]
    \centering
    \includegraphics[width=0.7\textwidth]{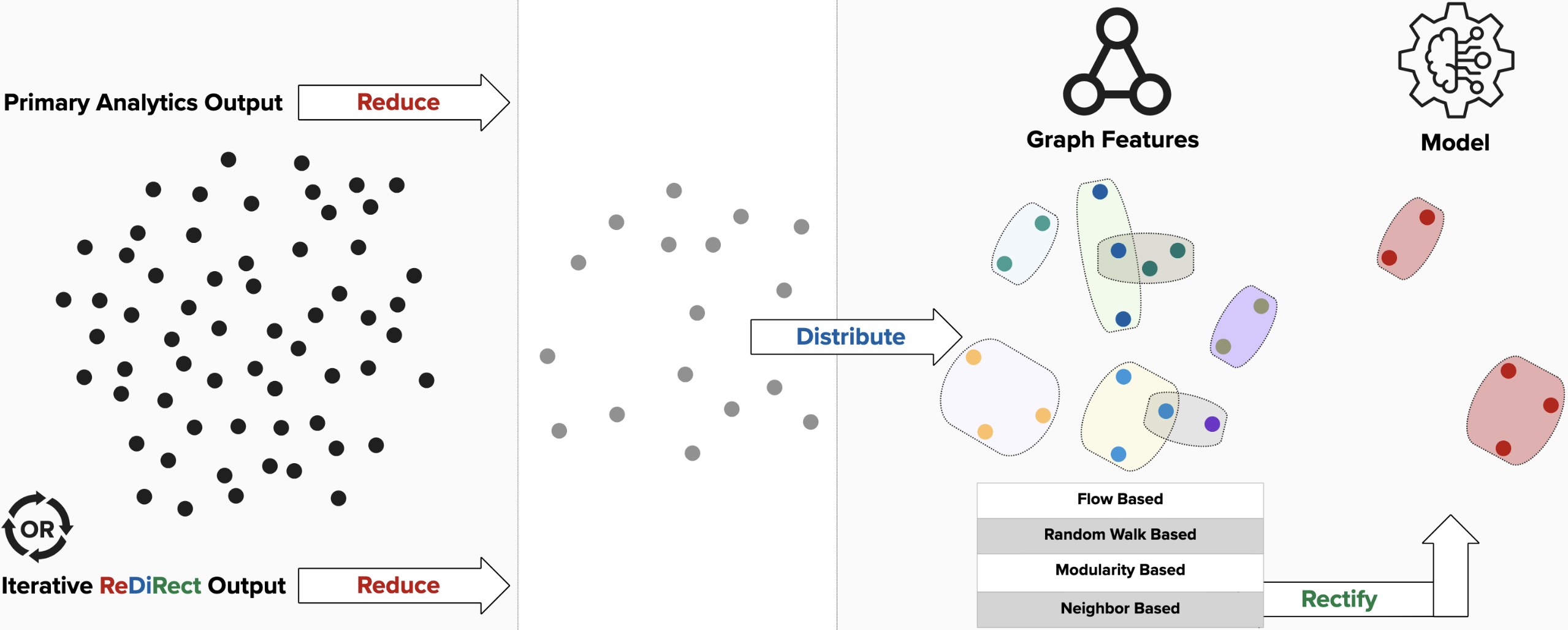}
    \caption{Conceptualization of the problem formulation for {\redirect}. Left side represents the actual data space; in the next part, nodes are filtered out, either by the \textit{primary} detection system; or by the output of an earlier {\redirect} run. Finally, the fuzzy communities (per node) are labeled as suspicious by an unsupervised machine learning model.}
    \label{fig:framework}
\end{figure*}

%As mentioned in Section \ref{sec:introduction}, 
We introduce a novel way of \textit{formulating} the AML problem as represented in Fig. \ref{fig:framework}.
%represents that formulation. 
In essence, we want to break down the problem into 3 parts, reduction of the search space (or \underline{Re}duce); distributed computation of complex graphs-based features (or \underline{Di}stribute); and (finally) identification and optimal \textit{filtering} of the money laundering patterns (or \underline{Rect}ify). We will now go through each part in detail.

% \vspace{-0.2cm}
\subsection{\texorpdfstring{\underline{Re}duce}{Reduce}}
The advantages of reducing the data scope are three-fold, which we will verify in Section \ref{sec:experiments} (Experiments),
(i) improved model accuracy,
(ii) reduction in ILT (Investigation Lead Time)\footnote{We define ILT as the mean time required by an AML analyst to close an alerted case of money laundering}, and, 
(iii) efficient generation of complex graph features.

The initial data space can be effectively reduced by applying simple risk-based AML rules. In practice, transaction monitoring systems in most financial institutions rely on extensive collections of such heuristics. For illustrative purposes, one might consider the following examples: flagging cash deposits exceeding a certain threshold; identifying accounts that perform a high number of rounded-amount deposits; or monitoring trading entities that send multiple transactions to high-risk jurisdictions within a short time window. However, incorporating such rules directly into a research context presents significant challenges. First, disclosing specific rule definitions could inadvertently inform potential money launderers, enabling them to circumvent detection by adapting their behavior. Second, the implementation of these rules in real banking environments typically relies on a broader and more detailed set of data attributes than those available in the synthetic datasets used for research purposes.
% \vspace{-0.1cm}
\subsubsection{\textbf{RM-1} (Reduction Method 1)}
For {$\mathcal{D}_{ibm}^{syn}$}, we instead reverse the problem by: first identifying the \textit{normal} entities, rather than identifying the suspicious entities using risk-based rules. We do so by using some common sense \textit{sequential} filtering, based on AML knowledge, which includes: 1) 
removing transactions with certain transaction types that can rarely be used in money laundering activities, 2)
removing transactions for accounts with a large number of transactions in a given period, and, 
3) removing transactions for accounts that interact with a large number of other accounts. The last two filters are indicators for legitimate businesses.
We want to reiterate that this is just a work-around. In a real-world setting, risk-based rules provide a much more practical approach with a much greater reduction of data space.
\subsubsection{\textbf{RM-2}}
For $\mathcal{D}_{lib}^{real}$, we employ more dynamic reduction methods (RM-2 and RM-3). For this method, we take the top X\% anomalous nodes produced by the first run of {\redirect} as the input data scope for the next run of {\redirect}. The idea here is that, after removing the most \textit{normal} nodes from the input data, we enable the model to identify the real anomalous nodes more effectively. The value for top X\% could be based on the risk appetite of the AML experts.

\subsubsection{\textbf{RM-3}}
This is kind of an extension of RM-2. For this method, we reduce the input data scope \textit{recursively} using the output of the previous {\redirect} run, as detailed in Algorithm \ref{alg:rec-red}.

\begin{algorithm}
	\caption{\textit{Recursive} (data scope) Reduction}
        \label{alg:rec-red}
	\begin{algorithmic}[1]
            \State initial\_size $\gets$ SIZE(accounts)
            \State reduce\_by $\gets$ 50\% (should be adjusted accordingly)
            \State break\_threshold $\gets$ 12\% (break if data scope < this \%)
            \State left\_percentage $\gets$ 100\%
		\While {true}
            \If{$left\_percentage$ < $break\_threshold$}
                \State \textbf{break}
            \EndIf
            \State anomalous $\gets$ {\redirect}(accounts)
            \State accounts $\gets$ TOP($reduce\_by$ \% $anomalous$)
            \State left\_percentage $\gets$ SIZE($accounts$) / $initial\_size$
		\EndWhile
	\end{algorithmic} 
\end{algorithm}

% \vspace{-0.2cm}
\subsection{\texorpdfstring{\underline{Di}stribute}{Distribute}}
From this point on, we \textit{assume} that the transactions produced by the last step are all alerts generated by the \textit{primary} (risk-based) transaction monitoring system of a bank. Our goal is now to reduce the huge volume of false positive alerts.
% \vspace{-0.1cm}
\subsubsection{Graph Generation}
From the datasets $\mathcal{D}_{lib}^{real}$ and $\mathcal{D}_{ibm}^{syn}$, the respective directed graphs can be extracted as $\mathcal{G = (V, E)}$. Where, $\mathcal{V}$ represents the bank accounts in the form of nodes (or vertices); and $\mathcal{E}$ represents the transactions in the form of edges. An aggregated (and unique) edge between $s, t \in \mathcal{V}$ is represented as $(s \rightarrow t) \in \mathcal{E}$. Where $s$ is the source account that sends the amount and $t$ is the target account that receives the amount. The total amount transferred from $s$ to $t$ is represented as an edge attribute, $\mathcal{A}_{s,t}$. The two node attributes, $\mathcal{S}$ and $\mathcal{R}$, represent the total amount sent and the total amount received by a node, respectively. The weight $\mathcal{W}$ for an edge is calculated as follows:
\begin{equation}
\label{eq:weight}
   \mathcal{W}(s \rightarrow t) = \frac{\mathcal{A}_{s \rightarrow t}}{\mathcal{S}_s} + \frac{\mathcal{A}_{s \rightarrow t}}{\mathcal{R}_t} 
\end{equation}

This weight serves as the main driver for constructing the communities described in the next section. The addition of the two ratios, one from the point of view of the sender and the other from the point of view of the receiver, ensures that a money launderer can not easily manipulate the system. For example, agents in a network can avoid appearing together (in a community) by using a high-volume intermediary account. In such a situation, only one of the two terms in Equation \ref{eq:weight} can be minimized, not both.
% \vspace{-0.1cm}
\subsubsection{Community Detection}
We employ the following methods to construct communities of nodes with close relationships:

\begin{itemize}
    \item (\textbf{CD-M}) Modularity-based community detection using the Leiden algorithm \cite{leiden}
    \item (\textbf{CD-RW}) Random walk-based subgraph detection\footnote{The distributed version of the algorithm can be found in the submitted code.} using the Personalized Page Rank algorithm \cite{pagerank}
\end{itemize}

\textbf{CD-RW} also serves as the \textit{context} for each alerted output. Using this context, an AML expert will assert the validity of an alert. The communities (or contexts) produced by this method have fuzzy boundaries. In other words, a node can belong to multiple communities. This aligns with the real-world scenario, where a money launderer can support multiple money laundering rings. For our main competing benchmark, \cite{ego2}, the same context is provided by the \textit{reduced ego-net} of each alerted account. The parameter $\theta$ (minimum allowed page rank) serves as the \textit{fuzziness} factor. It drives what constitutes a close relationship with respect to the \textit{central} node for which we are constructing a community. Consequently, it also limits the size of the resulting communities. The higher its value; the higher the overlap in communities. The value of $\theta$ can be adjusted according to risk appetite. This is also the parameter that drives the reduction of ILT.
% \vspace{-0.1cm}
\subsubsection{Feature Engineering}
Next, we generate features per \textbf{CD-M}; and per \textbf{CD-RW} community, in a distributed manner. The following are the main categories of features:
\begin{itemize}
    \item \textbf{Simple aggregations}: number of accounts and source-/target-only accounts; number of transactions; number of currencies; timestamp range and standard deviation.
    \item \textbf{Network metrics}: degree assortativity; diameter; maximum degree (in/out/all).
    \item \textbf{Turnover related}: total funds turnover (USD); currency-wise percentage contributions
\end{itemize}
In addition to these, we also generate the following two sets of node-level features:
\begin{itemize}
    \item \textbf{Flow-based}: funds transferred through the central node (up to \textit{n}-hops), depending on its role as a dispense, pass-through, or sink account
    \item \textbf{Neighbor-based}: aggregations and turnover-related features computed for incoming and outgoing neighbors
\end{itemize}

% \vspace{-0.1cm}
\subsection{\texorpdfstring{\underline{Rect}ify}{Rectify}}
In this part, we rectify the false positives in the output of the last {\redirect} run; or in the output of the \textit{primary} detection system, by correctly identifying the communities as anomalous. We train an IsolationForest model \cite{iforest} using the exhaustive set of features constructed in the \textit{\underline{Di}stribute} step. The anomaly score produced by IsolationForest can be used in combination with several risk-based rules to prioritize anomalies. Although we do not employ those rules here, again because of the risk implications.

If there are good quality labels available from the past, the training can also be done in a supervised learning setting. Or in a semi-supervised setting, where a combination of labeled and unlabeled data can be used to improve the quality of the predicted anomalies.

Optimal prioritization and filtering of anomalies is as crucial as identifying the correct anomalies. For the $\mathcal{D}_{ibm}^{syn}$ dataset, we employ a filtering method in which we iterate over the communities in descending order of their anomalousness. We keep a list of community members for each traversed community and remove those members from subsequent communities to ensure that an AML expert does not have to analyze the same \textit{context} over and over again.
% \vspace{-0.1cm}
\section{Experimental Evaluation}
\label{sec:experiments}
In this section, we will demonstrate the superiority of {\redirect} in terms of usefulness and scalability. The fully reproducible code for {\redirect}, as well as the experimental evaluation, is available here \footnote{https://github.com/mhaseebtariq/redirect}. All experiments were performed using a machine with the following specifications, chip: Apple M3 Pro; number of cores: 12; memory: 36 GB.

The (lack of) availability and the subjectivity that goes into money laundering labels make the problem space almost infeasible for supervised learning. We therefore build our framework in an unsupervised learning setting. In addition to that, we are designing our model in an entirely novel way. No existing work focuses on detecting the anomalous \textit{context} of a money laundering network; therefore, we cannot accurately compare the \textit{contextual} evaluation metric that we introduce in our experiments. The main focus of our evaluation will be on the accuracy (of the model); computational efficiency; and (reduction in) investigation lead times, compared to the benchmark methods. When evaluating our method against the $\mathcal{D}_{lib}^{real}$ dataset, we use the exact same metrics (TPR and TPR-AUC) and criteria as used in \cite{ego2}. In fact, Table \ref{table:libra-detailed-results} in Section \ref{sec:exp-results-libra} is an extension of Table 2 in \cite{ego2}. The thresholds used in \cite{ego2} to select the top anomalies are listed in Table \ref{table:libra-threshold}. We also used the exact thresholds for reporting the TPR's (True Positive Rates) for the $\mathcal{D}_{lib}^{real}$ dataset; and extrapolated them to define the thresholds for the $\mathcal{D}_{ibm}^{syn}$ dataset in Table \ref{table:ibm-threshold}.

\begin{table}[h]
% \small
% \setlength{\tabcolsep}{5pt}
% \renewcommand{\arraystretch}{1}
\caption{Thresholds used in \cite{ego2} to select the top anomalies. The top-x\% (predicted anomalies) is calculated from 385,100 accounts.}
\label{table:libra-threshold}
\begin{tabular}{|l|r|l|l|}
\hline
\textbf{Anomalies} & \textbf{Count} & \textbf{Threshold} & \textbf{\begin{tabular}[c]{@{}l@{}}Count / Actual\\ (Threshold Factor)\end{tabular}} \\ \hline
\textbf{Actual}        & \textbf{600}   & –    & \textbf{1.00}     \\ \hline
Top 0.1\%     & 385   & T1   & 0.64  \\ \hline
Top 0.2\%     & 770   & T2   & 1.28  \\ \hline
Top 0.5\%     & 1926  & T3   & 3.21  \\ \hline
Top 1\%       & 3851  & T4   & 6.42  \\ \hline
\end{tabular}
\end{table}

\begin{table}[h]
% \small
% \setlength{\tabcolsep}{5pt}
% \renewcommand{\arraystretch}{1}
\caption{Thresholds used in \cite{ego2} for $\mathcal{D}_{ibm}^{syn}$ \textit{large} datasets. LI = Low-Illicit; HI = High-Illicit.}
\label{table:ibm-threshold}
\begin{tabular}{|l|r|r|r|r|r|}
\hline
\textbf{Dataset} & \textbf{Total} & \textbf{T1 (0.64)} & \textbf{T2 (1.28)} & \textbf{T3 (3.21)} & \textbf{T4 (6.42)} \\ \hline
$\mathcal{D}_{lib}^{real}$                  & 600   & 385    & 770    & 1926   & 3851   \\ \hline
$\mathcal{D}_{ibm}^{syn}$ LI   & 1360  & 873    & 1745   & 4366   & 8729   \\ \hline
$\mathcal{D}_{ibm}^{syn}$ HI  & 9664  & 6201   & 12402  & 31021  & 62027  \\ \hline
\end{tabular}
\end{table}

\subsection{Experimental Setting}
The very definition of an anomaly could be subjective at best, incomplete, or outright wrong at worst. Whether the problem is modeled in a supervised, semi-supervised, self-supervised, or completely unsupervised setting, the starting point is always a set of assumptions. Assumptions are not facts and, therefore, are subjective. How can (then) those \textit{subjective} assumptions be \textit{objectively} evaluated? Identifying suspicious transactions, by isolating a suspiciously unique set of customer behaviors, is a rational design decision. Although not all suspicious transactions would be the result of those behaviors; neither all such behaviors would be indications of money laundering. Therefore, compared to other modeling problems, we argue that \textit{properly} validating anomalies is relatively more crucial in the context of AML modeling.

\subsubsection{Evaluation Metric}
\label{sec:eval-metric}
\begin{figure}[ht]
    \centering
    \includegraphics[width=1\linewidth]{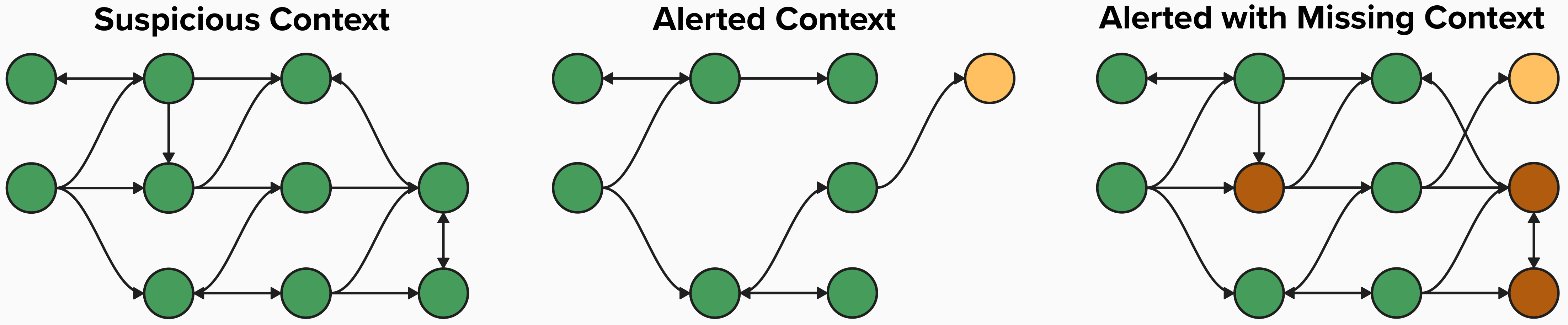}
    \caption{Example alerted flow: The yellow are extra (or false-positive); and the red are missing (or false-negative) nodes. The contextual completeness would be → 7 (correctly alerted context) / 11 (alerted with missing context) = 0.64.}
    \label{fig:cont_metric}
\end{figure}

Let $\mathcal{F}^r$ represent a real money laundering flow and $\mathcal{F}^d$, a flow detected as anomalous by {\redirect}. In the context of $\mathcal{D}_{ibm}^{syn}$, $\mathcal{F}^r$ are the injected patterns. Fig. \ref{fig:cont_metric} shows the simplified version of our proposed metric, applied to an example alerted flow.

For each $\mathcal{F}^r$ and $\mathcal{F}^d$, we have to identify how important it is in terms of compliance with the risk. This (again) is an area where we can employ risk-based rules to develop a comprehensive \textbf{risk scoring equation}. For our work, we are going to use the turnover magnitude as a proxy for that scoring. The rationale here is: the higher the turnover, the bigger the crime (must be) that produced those funds. Taking terms from Eq. \ref{eq:weight}, we define a flow turnover as:
\begin{equation}
\label{eq:turnover}
   \mathcal{T} = \sum_{i=1}^n \max((\mathcal{S}_i - \mathcal{R}_i), 0)
\end{equation}
Where, $n$ is the number of accounts in a flow. We further define the normalized version of $\mathcal{T}$ as:
\begin{equation}
\label{eq:turnover-nor}
   \thickbar{\mathcal{T}} = \min(\frac{\mathcal{T}}{\mathcal{C}}, 1)
\end{equation}

Where $\mathcal{C}$ is the capping factor, that we define based on the risk appetite. We can now also define the importance of each account in a flow. If in a money laundering network an account only has a minimal contribution in terms of its relative turnover, its exclusion from the network should not be penalized equally. We define the risk score for an account $i$ in $\mathcal{F}^r$ as follows:

\begin{equation}
\label{eq:turnover-acc}
   \mathcal{T}^a_i = \frac{(\mathcal{S}_i + \mathcal{R}_i)}{\mathcal{T}}
\end{equation}

The normalized version of Eq. \ref{eq:turnover-acc} can be defined as:

\begin{equation}
\label{eq:turnover-acc-nor}
   \thickbar{\mathcal{T}^a_i} = \frac{\mathcal{T}^a_i}{\sum_{j=1}^n \mathcal{T}^a_j}
\end{equation}

Using Eq. \ref{eq:turnover} and \ref{eq:turnover-acc-nor}, we can now construct our \textit{context-weighted} confusion matrix. Algorithm \ref{alg:wcm} describes the detailed implementation of the calculations. Once we have the context-weighted confusion matrix, the \textit{context-weighted} versions of the F-1 score; recall; precision, etc. can be inferred from the same numbers.

To evaluate the proposed \textit{context-weighted} recall in the context of AML modeling, we compare it against the normal recall metric. In Table \ref{table:ml-summary-combined}, the first column shows an example money laundering flow that we aim to detect using our model. For that flow, the third column shows the maximum money laundering contributions possible per account. Finally, the fourth and fifth columns show two scenarios for the detected context. In Scenario 1, a model identifies the anomalous context (as) involving accounts A, C, D, E, and F. In Scenario 2, a model identifies the context as A, B, and F. If we used the normal recall measure, we would have preferred the output (or detected context) of the first model. Although after a closer inspection, we can see that the second model captures the actors (most) \textit{central} to the money laundering activity; i.e., the source (A), as well as the destination (B) of the laundered funds.

\begin{table}[]
\caption{Summary of an example ML (money laundering) flow: transactions, maximum ML share (or contribution), and detection scenarios. Scenario 2 has better context-weighted recall than Scenario 1.}
\label{table:ml-summary-combined}
\begin{tabular}{|lll|l|l|}
\hline
\multicolumn{1}{|l|}{\textbf{\begin{tabular}[c]{@{}l@{}}Flow:\\ Source→Target\\ (Amount)\end{tabular}}} & \multicolumn{1}{l|}{\textbf{Node}} & \textbf{\begin{tabular}[c]{@{}l@{}}ML\\ Share\end{tabular}} & \textbf{\begin{tabular}[c]{@{}l@{}}Caught in\\ Scenario 1\end{tabular}} & \textbf{\begin{tabular}[c]{@{}l@{}}Caught in\\ Scenario 2\end{tabular}} \\ \hline
\multicolumn{1}{|l|}{F→A (5 K)}                                                                         & \multicolumn{1}{l|}{A}             & 11K                                                         & {\checkmark}                                                                       & {\checkmark}                                                                       \\ \hline
\multicolumn{1}{|l|}{A→D (1 K)}                                                                         & \multicolumn{1}{l|}{B}             & 10K                                                         & —                                                                       & {\checkmark}                                                                       \\ \hline
\multicolumn{1}{|l|}{C→A (2 K)}                                                                         & \multicolumn{1}{l|}{C}             & 2K                                                          & {\checkmark}                                                                       & {\checkmark}                                                                       \\ \hline
\multicolumn{1}{|l|}{E→A (2 K)}                                                                         & \multicolumn{1}{l|}{D}             & 1K                                                          & {\checkmark}                                                                       & —                                                                       \\ \hline
\multicolumn{1}{|l|}{A→B (10 K)}                                                                        & \multicolumn{1}{l|}{E}             & 2K                                                          & {\checkmark}                                                                       & —                                                                       \\ \hline
\multicolumn{1}{|l|}{—}                                                                                 & \multicolumn{1}{l|}{F}             & 5K                                                          & {\checkmark}                                                                       & —                                                                       \\ \hline
\multicolumn{3}{|r|}{\textbf{Recall}}                                                                                                                                                                      & \textbf{83\%}                                                                    & 50\%                                                                    \\ \hline
\multicolumn{3}{|r|}{\textbf{Context-weighted Recall}}                                                                                                                                                     & 68\%                                                                    & \textbf{84\%}                                                                    \\ \hline
\end{tabular}
\end{table}

\begin{algorithm}
	\caption{\textit{Context-Weighted} Confusion Matrix}
        \label{alg:wcm}
	\begin{algorithmic}[1]
            \State $\mathcal{X}$ $\leftarrow$ $\max$(‖$\mathcal{F}^d_1$‖, ‖$\mathcal{F}^d_2$‖, ..., ‖$\mathcal{F}^d_n$‖)
            \State SET weighted\_[TP,FP,TN,FN] $\leftarrow$ 0
		\For {$i$ in $\mathbfcal{F}^r$}
                \State $j$ $\leftarrow$ Get the index for the best matched $\mathcal{F}^r_i$ in $\mathbfcal{F}^d$
                \State matched $\leftarrow$ $\mathcal{F}^r_i$ $\cap$ $\mathcal{F}^d_j$
                \State TP $\leftarrow$ 0
                \For {$k$ in $matched$}
                    \State $TP$ $\mathrel{{+}{=}}$ $\thickbar{\mathcal{T}^a_k}$
                \EndFor
            \State $FP$ $\leftarrow$ ( (‖$\mathcal{F}^d_j$‖ - ‖$matched$‖ ) / ‖$\mathcal{F}^d_j$‖) $\times$ $\thickbar{\mathcal{T}_i}$
            \State $TN$ $\leftarrow$ ( ($\mathcal{X}$ - ‖$\mathcal{F}^d_j$‖) / $\mathcal{X}$ ) $\times$ $\thickbar{\mathcal{T}_i}$
            \State $FN$ $\leftarrow$ 1 - $TP$
            \If{$matched$ is empty}
                \State $TN$ $\leftarrow$ 0
                \State $FN$ $\leftarrow$ $\thickbar{\mathcal{T}_i}$
            \EndIf
            \State $weighted\_[*]$ += $[TP,FP,TN,FN]$
		\EndFor
            \State redundant\_flows $\leftarrow$ Flows in $\mathbfcal{F}^d$ with no match in $\mathbfcal{F}^r$
            \For {$l$ in $redundant\_flows$}
                \State $TP$, $FN$ $\leftarrow$ 0
                \State $FP$ $\leftarrow$ $\thickbar{\mathcal{T}_l}$
                \State $TN$ $\leftarrow$ ( ($\mathcal{X}$ - ‖$\mathcal{F}^d_l$‖) / $\mathcal{X}$ ) $\times$ $\thickbar{\mathcal{T}_l}$
                \State $weighted\_[*]$ += $[TP,FP,TN,FN]$
            \EndFor
	\end{algorithmic}
\end{algorithm}

\subsection{Results on the $\mathcal{D}_{lib}^{real}$ dataset}
\label{sec:exp-results-libra}

\begin{figure}[!tbp]
  \centering
  {\includegraphics[width=0.24\textwidth]{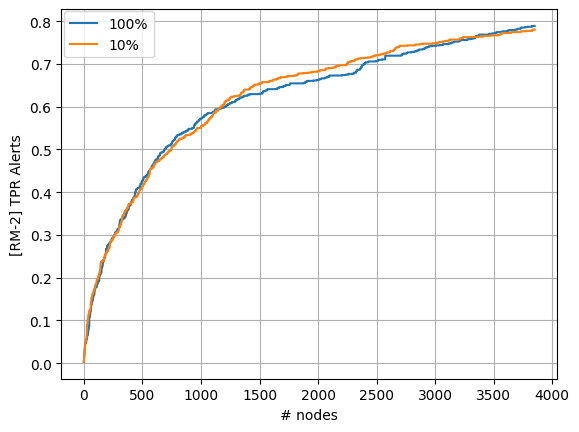}}
  \hfill
  % \vspace{-0.4cm}
  % \hspace{-1.4cm}
  {\includegraphics[width=0.24\textwidth]{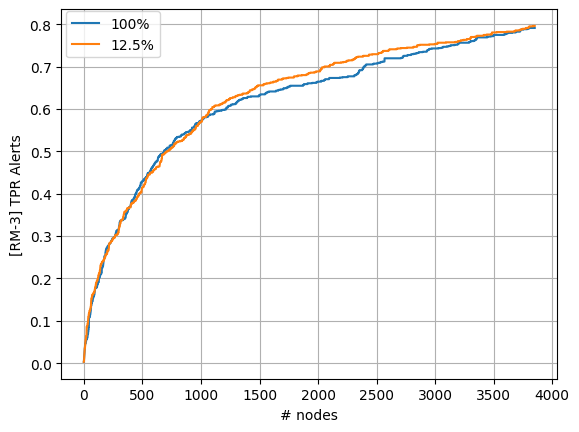}}
  \caption{TPR AUC plots comparing the initial run of {\redirect} (no reduction in data), with the outputs of RM-2 (10\% reduction) and RM-3 (12.5\% reduction) methods.}
  \label{fig:auc}
\end{figure}

Table \ref{table:libra-detailed-results} shows the clear superiority of {\redirect} in terms of the reported TPR's. In some cases, we see improvements of up to \textbf{$\sim$12\%}, even when the data scope is reduced to a very small proportion. In some cases, especially with higher selection thresholds, we see bigger improvements with greater reductions in the data scope. This claim can be further validated by analyzing Figure \ref{fig:auc}.

\subsubsection{Reduction in ILT (Investigation Lead Time)}
\label{sec:exp-results-ilt-red}
This is the main cost driver for any risk or AML department in a bank. Reducing the mean time it takes for an analyst to close an alert is of the utmost importance. To achieve that, we need to ensure that analysts have the most precise \textit{context} for each alert they are investigating. In the case of {\redirect}, that would be the n-hop community we construct for each node (or account). For \cite{ego2} that would be the reduced ego-net of each account.

For simplicity, we assume that an analyst takes a minute to investigate one account in an alerted \textit{context}. This is an ideal proxy for complexity, as a higher number of accounts translates to a higher number of interactions and network intricacies. We can therefore define ILT for a given alert as the size of the \textit{context} provided with that alert. Table \ref{table:ilt-red} shows the potential reduction in ILT, with decreasing size of the input data scope. In some cases, we have a reduction of up to a factor of \textbf{$\sim$6$\times$}. To put this into perspective, an analyst can now perform the duties of 6 analysts in the same amount of time.

\begin{table}[]
\caption{Mean investigation lead times. Bold font represents the best times. Cells with shaded background represent variations of \redirect.}
\label{table:ilt-red}
\begin{tabular}{|l|lllll|}
\hline
                                                           & \multicolumn{5}{l|}{\textbf{Mean ILT (Investigation Lead Time)}}                                                                                                        \\ \cline{2-6} 
\multirow{-2}{*}{\textbf{Method}}                          & \multicolumn{1}{l|}{}                & \multicolumn{4}{l|}{Threshold Factor}                                                                                            \\ \hline
\textbf{}                                                  & \multicolumn{1}{l|}{\textbf{Actual}} & \multicolumn{1}{l|}{\textbf{T1}}    & \multicolumn{1}{l|}{\textbf{T2}}    & \multicolumn{1}{l|}{\textbf{T3}}    & \textbf{T4}    \\ \hline
EgoNetRed \cite{ego2}                     & \multicolumn{1}{l|}{12.4}            & \multicolumn{1}{l|}{102.84}         & \multicolumn{1}{l|}{67.24}          & \multicolumn{1}{l|}{37.62}          & 23.59          \\ \hline
\cellcolor[HTML]{EFEFEF}\redirect 100\% & \multicolumn{1}{l|}{8.35}            & \multicolumn{1}{l|}{23.24}          & \multicolumn{1}{l|}{23.49}          & \multicolumn{1}{l|}{21.11}          & 17.54          \\ \hline
\cellcolor[HTML]{EFEFEF}{[}RM-2{]} 75\%                    & \multicolumn{1}{l|}{8.33}            & \multicolumn{1}{l|}{23.36}          & \multicolumn{1}{l|}{23.4}           & \multicolumn{1}{l|}{20.34}          & 16.59          \\ \hline
\cellcolor[HTML]{EFEFEF}{[}RM-2{]} 40\%                    & \multicolumn{1}{l|}{7.99}            & \multicolumn{1}{l|}{21.72}          & \multicolumn{1}{l|}{21.4}           & \multicolumn{1}{l|}{17.48}          & 13.63          \\ \hline
\cellcolor[HTML]{EFEFEF}{[}RM-2{]} 10\%                    & \multicolumn{1}{l|}{6.78}            & \multicolumn{1}{l|}{\textbf{16.67}} & \multicolumn{1}{l|}{\textbf{15.44}} & \multicolumn{1}{l|}{13.63}          & 10.98          \\ \hline
\cellcolor[HTML]{EFEFEF}{[}RM-3{]} 50\%                    & \multicolumn{1}{l|}{8.10}            & \multicolumn{1}{l|}{22.40}          & \multicolumn{1}{l|}{22.28}          & \multicolumn{1}{l|}{18.57}          & 14.38          \\ \hline
\cellcolor[HTML]{EFEFEF}{[}RM-3{]} 25\%                    & \multicolumn{1}{l|}{7.86}            & \multicolumn{1}{l|}{20.50}          & \multicolumn{1}{l|}{20.06}          & \multicolumn{1}{l|}{16.24}          & 12.93          \\ \hline
\cellcolor[HTML]{EFEFEF}{[}RM-3{]} 12.5\%                  & \multicolumn{1}{l|}{\textbf{6.47}}   & \multicolumn{1}{l|}{17.84}          & \multicolumn{1}{l|}{16.03}          & \multicolumn{1}{l|}{\textbf{13.08}} & \textbf{10.67} \\ \hline
\end{tabular}
\end{table}

\begin{table*}[tp!]
\caption{Detailed evaluation results on the $\mathcal{D}_{lib}^{real}$ dataset. \textbf{Bold} font represents the best version. \textcolor{Red}{Red} font is when {\redirect} is better than the competitors.}
\label{table:libra-detailed-results}
\begin{tabular}{|l|llllllllll}
\hline
{\color[HTML]{212529} }                         & \multicolumn{5}{l|}{{\color[HTML]{212529} Alerts}}                                                                                                                                                & \multicolumn{5}{l|}{{\color[HTML]{212529} Reports}}                                                                                                                                               \\ \cline{2-11} 
\multirow{-2}{*}{{\color[HTML]{212529} Method}} & \multicolumn{1}{l|}{AUC-\textbf{T4}}          & \multicolumn{1}{l|}{TPR-\textbf{T1}} & \multicolumn{1}{l|}{TPR-\textbf{T2}}          & \multicolumn{1}{l|}{TPR -\textbf{T3}}         & \multicolumn{1}{l|}{TPR -\textbf{T4}}         & \multicolumn{1}{l|}{AUC-\textbf{T4}}          & \multicolumn{1}{l|}{TPR-\textbf{T1}} & \multicolumn{1}{l|}{TPR-\textbf{T2}}          & \multicolumn{1}{l|}{TPR-\textbf{T3}}          & \multicolumn{1}{l|}{TPR-\textbf{T4}}          \\ \hline
\textbf{EgoNet \cite{oddball}}                                & 0.599                                  & 0.287                        & 0.437                                 & 0.660                                 & 0.815                                 & 0.574                                 & 0.146                        & 0.336                                 & 0.672                                 & 0.850                                   \\ \cline{1-1}
\textbf{EgoNetRed \cite{ego2}}                        & 0.604                                 & \textbf{0.405}               & 0.514                                 & 0.660                                 & 0.744                                 & 0.603                                 & \textbf{0.355}               & 0.437                                 & 0.696                                 & 0.772                                 \\ \cline{1-1}
\textbf{{\redirect} {100\%}}            & {\color[HTML]{CB0000} 0.617}          & 0.364                        & {\color[HTML]{CB0000} 0.524}          & {\color[HTML]{CB0000} 0.660}          & 0.791                                 & {\color[HTML]{CB0000} \textbf{0.676}} & 0.318                        & {\color[HTML]{CB0000} \textbf{0.546}} & {\color[HTML]{CB0000} \textbf{0.727}} & {\color[HTML]{CB0000} \textbf{0.955}} \\ \cline{1-1}
\textbf{\dittostraight {[}RM-2{] 75\%}}               & {\color[HTML]{CB0000} 0.623}          & 0.364                        & {\color[HTML]{CB0000} 0.527}          & {\color[HTML]{CB0000} 0.670}          & 0.804                                 & {\color[HTML]{CB0000} 0.674}          & 0.318                        & {\color[HTML]{CB0000} \textbf{0.546}} & {\color[HTML]{CB0000} \textbf{0.727}} & {\color[HTML]{CB0000} \textbf{0.955}} \\ \cline{1-1}
\textbf{{\dittostraight {[}RM-2{] 40\%}}}               & {\color[HTML]{CB0000} 0.634}          & 0.374                        & {\color[HTML]{CB0000} \textbf{0.529}} & {\color[HTML]{CB0000} 0.696}          & {\color[HTML]{CB0000} \textbf{0.819}} & {\color[HTML]{CB0000} 0.653}          & 0.318                        & 0.409                                 & {\color[HTML]{CB0000} \textbf{0.727}} & {\color[HTML]{CB0000} \textbf{0.955}} \\ \cline{1-1}
\textbf{{\dittostraight {[}RM-2{] 10\%}}}               & {\color[HTML]{CB0000} 0.623}          & 0.372                        & {\color[HTML]{CB0000} 0.510}          & {\color[HTML]{CB0000} 0.680}          & 0.781                                 & 0.586                                 & 0.318                        & 0.364                                 & 0.636                                 & 0.773                                 \\ \cline{1-1}
\textbf{{\dittostraight {[}RM-3{] 50\%}}}               & {\color[HTML]{CB0000} 0.631}          & 0.372                        & {\color[HTML]{CB0000} 0.525}          & {\color[HTML]{CB0000} 0.690}          & 0.814                                 & {\color[HTML]{CB0000} 0.658}          & 0.318                        & 0.409                                 & {\color[HTML]{CB0000} 0.727}          & {\color[HTML]{CB0000} 0.955}          \\ \cline{1-1}
\textbf{{\dittostraight {[}RM-3{] 25\%}}}               & {\color[HTML]{CB0000} 0.634}          & 0.369                        & {\color[HTML]{CB0000} 0.516}          & {\color[HTML]{CB0000} 0.700}          & 0.803                                 & {\color[HTML]{CB0000} 0.623}          & 0.318                        & 0.318                                 & {\color[HTML]{CB0000} 0.727}          & {\color[HTML]{CB0000} 0.909}          \\ \cline{1-1}
\textbf{{\dittostraight {[}RM-3{] 12.5\%}}}             & {\color[HTML]{CB0000} 0.629}          & 0.368                        & {\color[HTML]{CB0000} 0.513}          & {\color[HTML]{CB0000} 0.685}          & 0.796                                 & 0.593                                 & 0.318                        & 0.318                                 & 0.636                                 & 0.773                                 \\ \cline{1-1}
\end{tabular}
\end{table*}

\subsection{Results on the $\mathcal{D}_{ibm}^{syn}$ dataset}
\label{sec:results-results-libra}
Here, we are going to focus on the \textit{context-weighted} recall metric (/confusion matrix) defined earlier. We want to stress here that the same metric cannot be applied to the ${D}_{lib}^{real}$ dataset, or any other openly available dataset out there. Only because of the \textit{rich} flow-level labels present in $\mathcal{D}_{ibm}^{syn}$, we are able to apply our proposed metric.

Table \ref{table:ibm-recall} reports the \textit{context-weighted} recall with different data reduction methods. It is again clear from the results that the reduction in the input data scope considerably improves accuracy. However, RM-2 does not work well here. We believe that this is due to the presence of an \textit{unrealistically} high number of accounts involved in money laundering activities. For such a dataset, a heuristic-based approach (like RM-1) might be more suitable. Using Algorithm \ref{alg:wcm} we can also construct the \textit{context-weighted} confusion matrix as shown in Table \ref{table:conmat}, for a more holistic view of the model performance. An interesting thing to note is that the numbers reported here are \textit{even} (slightly) better than those produced from \cite{graphfep}, where the model is trained in a \textbf{supervised} setting instead. Although, to state the obvious, the comparison is not entirely fair, as we have different data scope and a different variation of metrics.

\begin{table}[]
\caption{\textit{Context-weighted} recall for the $\mathcal{D}_{ibm}^{syn}$ datasets. LI = Low-Illicit; HI = High-Illicit.}
\label{table:ibm-recall}
\begin{tabular}{|l|llll|}
\hline
\multirow{2}{*}{Method}        & \multicolumn{4}{l|}{Threshold Factor}                                                                                \\ \cline{2-5} 
                               & \multicolumn{1}{l|}{\textbf{T1}} & \multicolumn{1}{l|}{\textbf{T2}} & \multicolumn{1}{l|}{\textbf{T3}} & \textbf{T4} \\ \hline
\textbf{{[}LI{]} RM-1}         & \multicolumn{1}{l|}{0.034}       & \multicolumn{1}{l|}{0.075}       & \multicolumn{1}{l|}{0.190}       & 0.351       \\ \hline
\textbf{{[}HI{]} No reduction} & \multicolumn{1}{l|}{0.050}       & \multicolumn{1}{l|}{0.087}       & \multicolumn{1}{l|}{0.271}       & 0.450       \\ \hline
\textbf{{[}HI{]} RM-1}         & \multicolumn{1}{l|}{0.287}       & \multicolumn{1}{l|}{0.498}       & \multicolumn{1}{l|}{0.675}       & 0.713       \\ \hline
\textbf{{[}HI{]} RM-2 50\%}    & \multicolumn{1}{l|}{0.237}       & \multicolumn{1}{l|}{0.477}       & \multicolumn{1}{l|}{0.658}       & 0.683       \\ \hline
\end{tabular}
\end{table}

\begin{table}[h]
% \small
% \setlength{\tabcolsep}{5pt}
% \renewcommand{\arraystretch}{1}
\caption{\textit{Context-weighted} confusion matrix for $\mathcal{D}_{ibm}^{syn}$ (high-illicit) [RM-1], with T4 selection threshold.}
\label{table:conmat}
\centering
\begin{tabular}{|c|c|r|r|r|}
\hline
\multicolumn{2}{|c|}{\textbf{}}           & \textbf{p} & \textbf{n} & \textbf{Total}     \\ \hline
\multirow{2}{*}{\textbf{Actual}} 
& \textbf{p'}     & 369          & 149          & \textbf{518}   \\ \cline{2-5}
& \textbf{n'}     & 3497         & 63063        & \textbf{66560} \\ \hline
\multicolumn{2}{|c|}{\textbf{Total}}      & \textbf{3866} & \textbf{63211} &                   \\ \hline
\end{tabular}
\end{table}

\subsection{Execution times}
\label{sec:exp-exec-times}
In terms of scalability, we were able to run all of our experiments on a personal laptop. Even executions for the larger $\mathcal{D}_{ibm}^{syn}$ datasets, with up to 180 million transactions, do not take more than a couple of hours.

\begin{figure}[ht]
    \centering
    \includegraphics[width=0.8\linewidth]{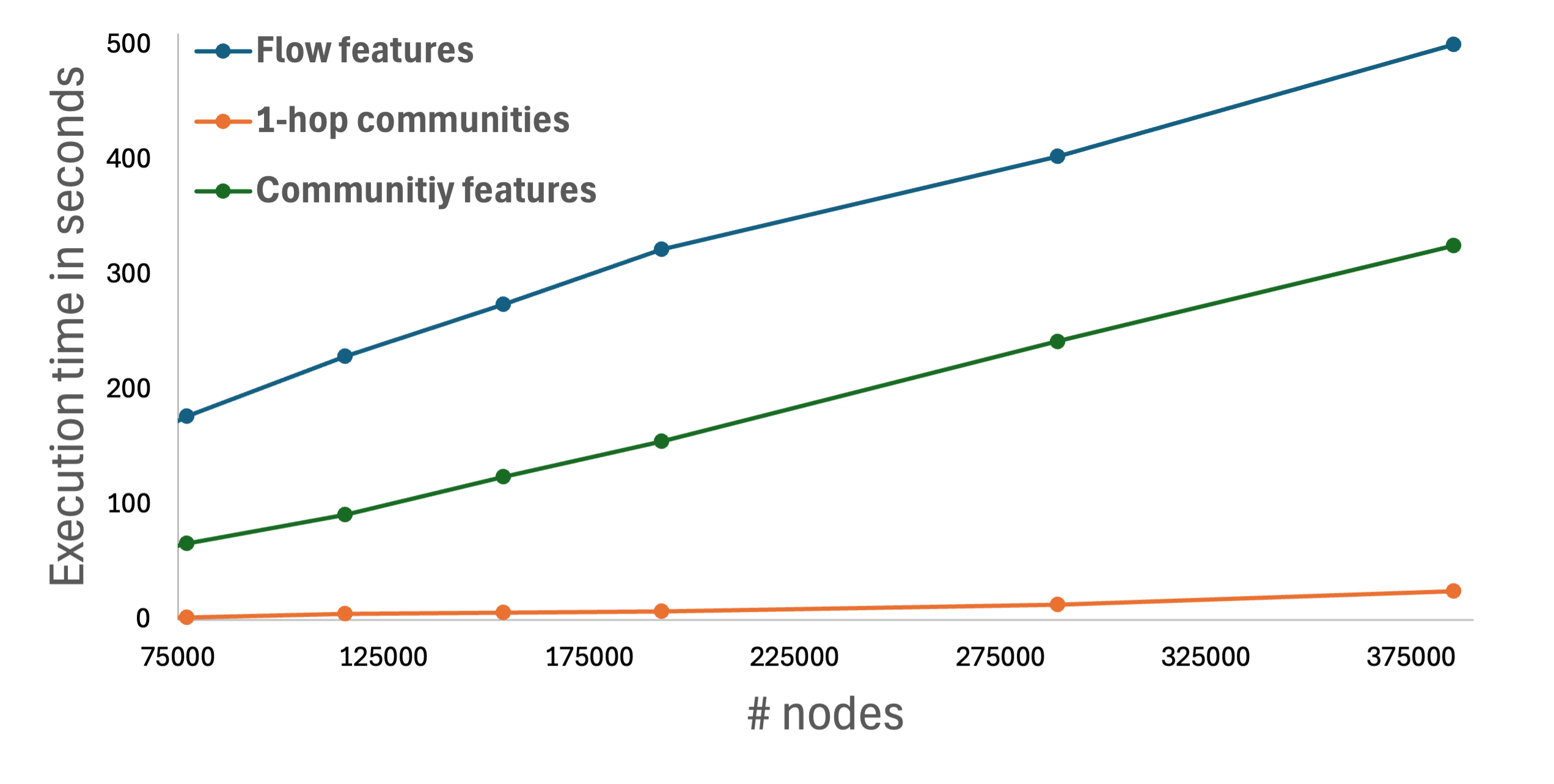}
    \caption{The execution times for each of the heavy duty tasks in the framework, with increasing number of nodes in the $\mathcal{D}_{lib}^{real}$ dataset.}
    \label{fig:exec-times}
\end{figure}

\begin{figure}
    \centering
    \includegraphics[width=0.8\linewidth]{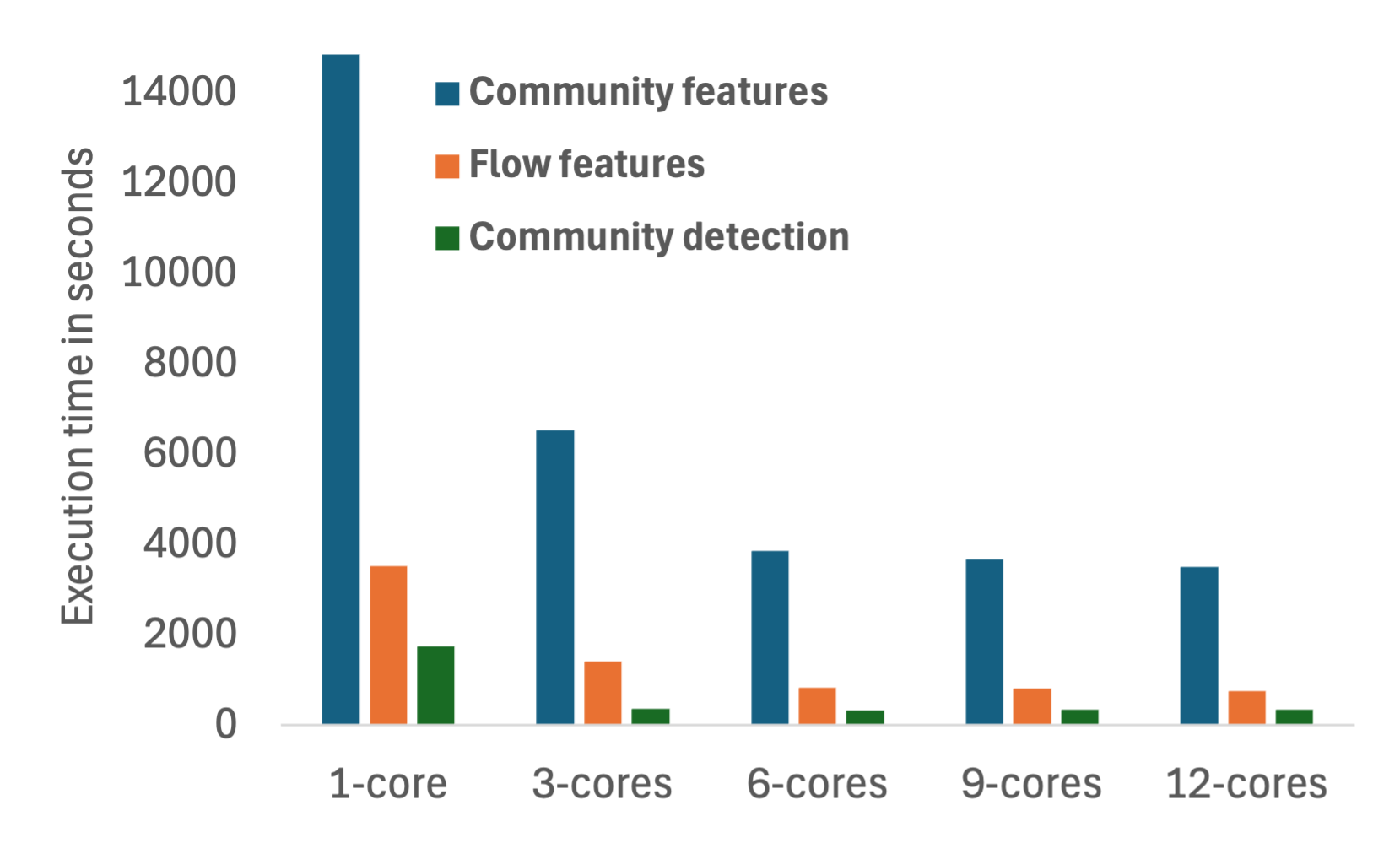}
    \caption{Execution times for $\mathcal{D}_{ibm}^{syn}$ large dataset, with increasing level of distribution.}
    \label{fig:dist-stats}
\end{figure}

From Figure \ref{fig:exec-times}, we can validate that the framework scales almost linearly with increasing data size. Furthermore, Figure \ref{fig:dist-stats} shows how the execution times for different steps can be reduced (drastically) by having higher levels of parallelization.

\section{Conclusion and Future Work}
\label{sec:conclusion}
We showed that with {\redirect} it is possible to generate high-quality and \textit{comprehensive} money laundering alerts, while keeping false positives to a minimum. Not just the overall false positive signals; but also false positives within a community identified as anomalous. The fine tuning of the last step \textit{\underline{Rect}ify} has much room for improvement. We can employ stratified and stochastic model trainings to minimize the impact of groups of nodes appearing in multiple communities. Advanced deep learning models such as GNN \cite{gnn}, autoencoders \cite{autoencoder, vgae}, etc. could be suitable candidates to identify anomalous communities in a more robust manner.

The selection of the top anomalies is also an area where we can have some improvements. Anomalous communities with overlapping nodes can be joined or merged in a smart way, reducing (even further) the number of anomalies to analyze. We can also apply AML typology-specific risk scoring to shortlist more relevant alerts. For example, to capture trade-based money laundering \cite{tbml} networks, we can select communities with (certain types of) businesses that are mostly used in such schemes. We plan to cover all of these areas of improvement in our future research. Finally, we invite fellow-researchers to use Tables \ref{table:ibm-recall} and \ref{table:conmat} as the first benchmarks, based on the (proposed) \textit{context-weighted} metrics.

\printbibliography

@article{flowscope, title={FlowScope: Spotting Money Laundering Based on Graphs}, volume={34}, abstractNote={&lt;p&gt;Given a graph of the money transfers between accounts of a bank, how can we detect money laundering? Money laundering refers to criminals using the bank’s services to move massive amounts of illegal money to untraceable destination accounts, in order to inject their illegal money into the legitimate financial system. Existing graph fraud detection approaches focus on dense subgraph detection, without considering the fact that money laundering involves high-volume &lt;em&gt;flows&lt;/em&gt; of funds through chains of bank accounts, thereby decreasing their detection accuracy. Instead, we propose to model the transactions using a multipartite graph, and detect the complete flow of money from source to destination using a scalable algorithm, FlowScope. Theoretical analysis shows that FlowScope provides guarantees in terms of the amount of money that fraudsters can transfer without being detected. FlowScope outperforms state-of-the-art baselines in accurately detecting the accounts involved in money laundering, in both injected and real-world data settings.&lt;/p&gt;}, number={04}, journal={Proceedings of AAAI}, author={Li, Xiangfeng and Liu, Shenghua and Li, Zifeng and Han, Xiaotian and Shi, Chuan and Hooi, Bryan and Huang, He and Cheng, Xueqi}, year={2020}, month={04}, pages={4731-4738} }

@inproceedings{graphfep,
author = {Blanu\v{s}a, Jovan and Cravero Baraja, Maximo and Anghel, Andreea and von Niederh\"{a}usern, Luc and Altman, Erik and Pozidis, Haris and Atasu, Kubilay},
title = {Graph Feature Preprocessor: Real-time Subgraph-based Feature Extraction for Financial Crime Detection},
numpages = {9},
series = {ICAIF '24}
}

@inproceedings{synthdata,
author = {Altman, Erik and Blanu\v{s}a, Jovan and von Niederh\"{a}usern, Luc and Egressy, B\'{e}ni and Anghel, Andreea and Atasu, Kubilay},
title = {Realistic synthetic financial transactions for anti-money laundering models},
booktitle = {NIPS '23},
articleno = {1300},
numpages = {24}
}

@article{smurfing,
 author = {Welling, S.N.},
 journal = {Fla. Law},
 pages = {287--343},
 title = {Smurfs, money laundering and the federal criminal law: The crime of structuring transactions},
 volume = {41},
 year = {1989}
}

@online{mule,
    author = "Europol",
    title = "Money Muling",
    url  = "https://www.europol.europa.eu/operations-services-and-innovation/public-awareness-and-prevention-guides/money-muling",
    addendum = "(accessed: 14.06.2023)"
}

@online{mloverview,
    author = "UNODC",
    title = "Money Laundering Overview",
    url  = "https://www.unodc.org/unodc/en/money-laundering/overview.html",
    addendum = "(accessed: 25.03.2023)"
}

@online{amlstats,
    author = "Doug Bonderud",
    title = "AML statistics of 2023",
    url  = "https://withpersona.com/blog/the-most-mind-blowing-money-laundering-statistics-of-2022",
    addendum = "(accessed: 08.03.2025)"
}

@article{study,
author = {Zhao, Zhongying and Zheng, Shaoqiang and Li, Chao and Sun, Jinqing and Chang, Liang and Chiclana, Francisco},
year = {2018},
month = {06},
title = {A comparative study on community detection methods in complex networks},
volume = {35},
journal = {Journal of Intelligent \& Fuzzy Systems}
}

@article{leiden,
author={Traag, V. A. and Waltman, L. and van Eck, N. J.},
title={From Louvain to Leiden: guaranteeing well-connected communities},
journal={Scientific Reports},
year={2019},
volume={9},
number={1},
pages={5233}
}

@article{Blondel_2008,
	year = 2008,
	month = {10},
	publisher = {{IOP} Publishing},
	volume = {2008},
	number = {10},
	pages = {P10008},
	author = {Vincent D Blondel and Jean-Loup Guillaume and Renaud Lambiotte and Etienne Lefebvre},
	title = {Fast unfolding of communities in large networks},
	journal = {Journal of Statistical Mechanics: Theory and Experiment}
}

@misc{wagenseller2017size,
  author={Wagenseller, Paul et al.},
  journal={IEEE Transactions on Computational Social Systems}, 
  title={Size Matters: A Comparative Analysis of Community Detection Algorithms}, 
  year={2018},
  volume={5},
  number={4},
  pages={951-960}
}

@misc{akoglu2014graphbased,
author={Akoglu, Leman et al.},
title={Graph based anomaly detection and description: a survey},
journal={Data Mining and Knowledge Discovery},
year={2015},
doi = {10.1007/s10618-014-0365-y}
}

@inproceedings{lime,
author = {Ribeiro, Marco Tulio et al.},
title = {"Why Should I Trust You?": Explaining the Predictions of Any Classifier},
year = {2016},
doi = {10.1145/2939672.2939778},
booktitle = {Proceedings of the 22nd ACM SIGKDD},
numpages = {10},
}

@article{shap,
  title={From local explanations to global understanding with explainable AI for trees},
  author={Lundberg, Scott M. and Erion, Gabriel and Chen, Hugh and DeGrave, Alex and Prutkin, Jordan M. and Nair, Bala and Katz, Ronit and Himmelfarb, Jonathan and Bansal, Nisha and Lee, Su-In},
  journal={Nature Machine Intelligence},
  volume={2},
  number={1},
  pages={2522-5839},
  year={2020},
  publisher={Nature Publishing Group}
}

@misc{vgae,
      title={Variational Graph Auto-Encoders}, 
      author={Thomas N. Kipf et al.},
      year={2016},
      url={https://arxiv.org/abs/1611.07308}, 
}

@inproceedings{graphsage,
author = {Hamilton, William L. et al.},
title = {Inductive representation learning on large graphs},
year = {2017},
isbn = {9781510860964},
booktitle = {Proceedings of the 31st NIPS},
numpages = {11},
}

@article{supervised,
author = {David Savage and Qingmai Wang and Pauline Lienhua Chou and Xiuzhen Zhang and Xinghuo Yu},
journal={ArXiv},
year = {2016},
month = {08},
title = {Detection of money laundering groups using supervised learning in networks}
}

@InProceedings{oddball,
author="Akoglu, Leman et al.",
title="oddball: Spotting Anomalies in Weighted Graphs",
booktitle="Advances in Knowledge Discovery and Data Mining",
year="2010",
isbn="978-3-642-13672-6"
}

@ARTICLE{ego2,
  author={Dumitrescu, Bogdan et al.},
  journal={IEEE Access}, 
  title={Anomaly Detection in Graphs of Bank Transactions for Anti Money Laundering Applications}, 
  year={2022},
  volume={10},
  number={},
  doi={10.1109/ACCESS.2022.3170467}
}

@article{modularity,
author = {M. E. J. Newman },
title = {Modularity and community structure in networks},
journal = {Proceedings of the National Academy of Sciences},
volume = {103},
number = {23},
pages = {8577-8582},
year = {2006}}

@InProceedings{exstraqt,
author="Tariq, Haseeb and Hassani, Marwan",
title="Extracting Money Laundering Transactions from Quasi-Temporal Graph Representation",
booktitle="ACM SIGAPP SAC '26",
doi={10.1145/3748522.3779790},
}

@InProceedings{fastman,
author="Tariq, Haseeb and Hassani, Marwan",
title="Topology-Agnostic Detection of Temporal Money Laundering Flows in Billion-Scale Transactions",
booktitle="PKDD '25",
isbn="978-3-031-74643-7",
doi={10.1007/978-3-031-74643-7_29},
}

@techreport{pagerank,
           month = {11},
          author = {Lawrence Page et al.},
           title = {The PageRank Citation Ranking: Bringing Order to the Web.},
            type = {Technical Report},
       publisher = {Stanford InfoLab},
            year = {1999},
     institution = {Stanford InfoLab}
}

@inproceedings{cubeflow,
author = {Sun, Xiaobing and Zhang, Jiabao and Zhao, Qiming and Liu, Shenghua and Chen, Jinglei and Zhuang, Ruoyu and Shen, Huawei and Cheng, Xueqi},
title = {CubeFlow: Money Laundering Detection with Coupled Tensors},
booktitle = {PAKDD 2021},
numpages = {13}
}

@inproceedings{monlad,
author = {Sun, Xiaobing and Feng, Wenjie and Liu, Shenghua and Xie, Yuyang and Bhatia, Siddharth and Hooi, Bryan and Wang, Wenhan and Cheng, Xueqi},
title = {MonLAD: Money Laundering Agents Detection in Transaction Streams},
year = {2022},
booktitle = {WSDM '22},
pages = {976–986},
numpages = {11}
}

@inproceedings{fastppr,
author = {Lofgren, Peter A. and Banerjee, Siddhartha and Goel, Ashish and Seshadhri, C.},
title = {FAST-PPR: scaling personalized pagerank estimation for large graphs},
year = {2014},
booktitle = {Proceedings of SIGKDD },
pages = {1436–1445},
numpages = {10}
}

@online{abnfine,
    author = "Reuters",
    title = "ABN Amro to settle money laundering probe for \$574 mln",
    url  = "https://www.reuters.com/business/abn-amro-settle-money-laundering-probe-574-million-2021-04-19/",
    addendum = "(accessed: 14.03.2023)"
}

@INPROCEEDINGS{iforest,
  author={Liu, Fei et al.},
  booktitle={2008 Eighth IEEE International Conference on Data Mining}, 
  title={Isolation Forest}, 
  volume={},
  number={},
  pages={413-422}}

@article{gnn,
  author={Scarselli, Franco and Gori, Marco and Tsoi, Ah Chung and Hagenbuchner, Markus and Monfardini, Gabriele},
  journal={IEEE Transactions on Neural Networks}, 
  title={The Graph Neural Network Model}, 
  year={2009},
  volume={20},
  number={1},
  pages={61-80}}

@InProceedings{autoencoder,
  title =    {Autoencoders, Unsupervised Learning, and Deep Architectures},
  author =   {Baldi, Pierre},
  booktitle =    {Proceedings of ICML},
  pages =    {37--49},
  year =     {2012},
  volume =   {27},
  month =    {7},
  publisher =    {PMLR}
}

@online{tbml,
    author = "FATF",
    title = "Trade-Based Money Laundering",
    url  = "http://fatf-gafi.org/en/publications/Methodsandtrends/Trade-basedmoneylaundering.html",
    addendum = "(accessed: 15.03.2023)"
}

@misc{pet,
      title={Privacy-Enhancing Technologies for Financial Data Sharing}, 
      author={Panagiotis Chatzigiannis and Wanyun Catherine Gu and Srinivasan Raghuraman and Peter Rindal and Mahdi Zamani},
      year={2023},
}

@article{amlgraph,
author = {Karim, Rezaul et al.},
title = {Scalable Semi-Supervised Graph Learning Techniques for Anti Money Laundering},
doi = {10.1109/ACCESS.2024.3383784}
}

\end{document}